\documentclass[dvipsnames,format=sigconf,anonymous=False,review=False]{acmart}

\usepackage{booktabs}
\usepackage{siunitx}
\usepackage{subfigure}

\AtBeginDocument{%
  \providecommand\BibTeX{{%
    \normalfont B\kern-0.5em{\scshape i\kern-0.25em b}\kern-0.8em\TeX}}}

\copyrightyear{2023}
\acmYear{2023}
\setcopyright{rightsretained}
\acmConference[GECCO '23 Companion]{Genetic and Evolutionary Computation Conference Companion}{July 15--19, 2023}{Lisbon, Portugal}
\acmBooktitle{Genetic and Evolutionary Computation Conference Companion (GECCO '23 Companion), July 15--19, 2023, Lisbon, Portugal}\acmDOI{10.1145/3583133.3590601}
\acmISBN{979-8-4007-0120-7/23/07}

\begin{document}

\title{Comparison of Single- and Multi- Objective Optimization Quality for Evolutionary Equation Discovery}


\author{Mikhail Maslyaev}
\email{maslyaitis@gmail.com}
\affiliation{%
  \institution{ITMO University}
  \streetaddress{Kronverkskiy Prospekt, 49}
  \city{St Petersburg}
  \country{Russia}
  \postcode{197101}
}

\author{Alexander Hvatov}
\email{alex_hvatov@itmo.ru}
\affiliation{%
  \institution{ITMO University}
  \streetaddress{Kronverkskiy Prospekt, 49}
  \city{St Petersburg}
  \country{Russia}
  \postcode{197101}
}

\renewcommand{\shortauthors}{M. Maslyaev, and A. Hvatov}

\begin{abstract}
Evolutionary differential equation discovery proved to be a tool to obtain equations with less a priori assumptions than conventional approaches, such as sparse symbolic regression over the complete possible terms library. The equation discovery field contains two independent directions. The first one is purely mathematical and concerns differentiation, the object of optimization and its relation to the functional spaces and others. The second one is dedicated purely to the optimizatioal problem statement. Both topics are worth investigating to improve the algorithm's ability to handle experimental data a more artificial intelligence way, without significant pre-processing and a priori knowledge of their nature. In the paper, we consider the prevalence of either single-objective optimization, which considers only the discrepancy between selected terms in the equation, or multi-objective optimization, which additionally takes into account the complexity of the obtained equation. The proposed comparison approach is shown on classical model examples -- Burgers equation, wave equation, and Korteweg - de Vries equation.
\end{abstract}



\maketitle
\section{Introduction}
The recent development of artificial intelligence has given high importance to problems of interpretable machine learning. In many cases, users value models not only for their quality of predicting the state of the studied system but also for the ability to provide some information about its operation. In the case of modeling physical processes, commonly, the most suitable models have forms of partial differential equations. Thus many recent studies aimed to develop the concept of data-driven differential equations discovery. In the paper, data-driven discovery implies obtaining a differential equation from a set of empirical measurements, describing the dynamics of a dependent variable in some domain. Furthermore, equation-based models can be incorporated into pipelines of automated machine learning, that can include arbitrary submodels, with approach, discussed in paper \cite{sarafanov2022evolutionary}.


Initial advances in differential equations discovery were made with symbolic regression algorithm, as in \cite{symb_reg_ODE}. The algorithm employs genetic programming to detect the graph, that represents differential equation. 
One of the groups of the most simple yet practical techniques of equation construction is based on the sparse linear regression (least absolute shrinkage and selection operator), introduced in works \cite{messenger2021weak}, \cite{schaeffer2017learning}, \cite{schaeffer_etal2017learning}, and other similar projects. 
This approach has limited flexibility, having applicability restrictions in cases of the equation with low magnitude coefficients, being discovered on noisy data. This issue is addressed by employing Bayesian interference as in \cite{LASSO_Bootstrap} to estimate the coefficients of the equation, as in work \cite{gao2023convergence}. To account for the uncertainty in the resulting model, the approximating term library can be biased statistically \cite{fasel2022ensemble}.
Physics-informed neural networks (PINN) form the next class of data-driven equation discovery tools, representing the process dynamics with artificial neural networks. 
The primary research on this topic is done in work \cite{pinn}, while recent advances have been made in incorporating more complex types of neural networks in the PINNs \cite{gao2022physics, Thanasutives_2023}.

In recent studies \cite{maslyaev2021partial, lu2021deepxde}, evolutionary algorithms have proved to be a rather flexible tool for differential equation discovery, demanding only a few assumptions about the process properties. The problem is stated as the process representation error minimization. 
Implementing multi-objective evolutionary optimization, first introduced for DE systems, as in \cite{CEC_EPDE}, seems to be a feasible way to improve the quality of the equation search, operating on fewer initial assumptions and providing higher diversity among the processed candidates. Additional criteria can represent other valuable properties of the constructed models, namely conciseness.

This study compares the performance of single- and multi- objective optimization. Namely, the hypothesis that the multi-objective optimization creates and preserves diversity in the population and thus may achieve a better fitness function values, than that of a single-objective approach.
The theoretical comparison shows that multi-objective algorithms allow escaping local minima as soon as the number of objectives is reasonably small \cite{ishibuchi2006comparison}. For equation discovery applications, the function landscapes have a more complex structure, so increased diversity of the population can benefit the resulting quality.

\section{Algorithm description}
\label{sec:alg_descr}

The data-driven differential equation identification operates on problems of selecting a model for dynamics of the variable $u = u(t, \mathbf{x})$ in a spatio-temporal domain $(0, T) \bigtimes \Omega$, that is implicitly described by differential equation Eq.~\ref{eq:problem_statement} with corresponding initial and boundary conditions. It can be assumed, that the order of the unknown equation can be arbitrary, but rather low (usually of second or third order). 

\begin{equation}
   F(t, \mathbf{x}, u, \frac{\partial u}{\partial t}, \frac{\partial u}{\partial x_1}, \; ... \; \frac{\partial u}{\partial x_n}) = 0
\label{eq:problem_statement}
\end{equation}

Both multi-objective and single-objective approaches have the same core of "graph-like" representation of a differential equation (encoding) and similar evolutionary operators that will be described further.

\subsection{Differential equation representation}
\label{subsec:representation}

To represent the candidate differential equation the computational graph structure is employed. A fixed three-layer graph structure is employed to avoid the infeasible structures, linked to unconstrained graph construction and overtraining issues, present in symbolic regression. The lowest level nodes contain tokens, middle nodes and the root are multiplication and summation operations. 
The data-driven equations take the form of a linear combination of product terms, represented by the multiplication of derivatives, other functions and a real-valued coefficient Eq.~\ref{eq:data_driven_statement}. 

\begin{equation}
    \begin{cases}
          {F'}(t, \mathbf{x}, u, \frac{\partial u}{\partial t}, \frac{\partial u}{\partial x_1}, \; ... \; \frac{\partial u}{\partial x_n}) = \sum_{i} \alpha_{i} \prod_{j} f_{ij} = 0 \\
          {G'}(u) |_{\Gamma} = 0 
    \end{cases}
\label{eq:data_driven_statement}
\end{equation}

Here, the factors $f_{ij}$ are selected from the user-defined set of elementary functions, named tokens. The problem of an equation search transforms into the task of detecting an optimal set of tokens to represent the dynamics of the variable $u(t, \mathbf{x})$, and forming the equation by evaluating the coefficients $\mathbf{\alpha} = (\alpha_1, \; ... \; \alpha_m)$.


During the equation search, we operate with tensors of token values, evaluated on grids $u_{\gamma} = u(t_{\gamma}, \mathbf{x}_{\gamma})$ in the processed domain $(0, T) \bigtimes \Omega$. 


Sparsity promotion in the equation operates by filtering out nominal terms with low predicting power and is implemented with LASSO regression. For each individual, a term (without loss of generality, we can assume that it is the $m$-th term) is marked to be a "right-hand side of the equation" for the purposes of term filtering and coefficient calculation. The terms $T_i = \prod_{j} f_{ij}$ are paired with real-value coefficients obtained from the optimization subproblem of Eq.~\ref{eq:LASSO}. Finally, the equation coefficients are detected by linear regression. 

\begin{equation}
    \mathbf{\alpha'} = \text{arg}\min \limits_\alpha (|| \sum_{i, \; i \neq m} \alpha'_{i} \prod_{j} f_{ij} - \prod_{j} f_{mj}||_2 + \lambda || \mathbf{\alpha'} ||_1)
    \label{eq:LASSO}
\end{equation}


In the initialization of the algorithm equation graphs are randomly constructed for each individual from the sets of user-defined tokens with a number of assumptions about the structures of the ``plausible equations''. 

\subsection{Mechanics of implemented evolutionary operators}
\label{subsec:evo_operators}

To direct the search for the optimal equations, standard evolutionary operators of mutation and cross-over have been implemented. While the mechanics of single- and multi-objective optimization in the algorithm differ, they work similarly on the stage of applying equation structure-changing operators. With the graph-like encoding of candidate equations, the operators can be represented as changes, introduced into its subgraphs.

The algorithm properties to explore structures are provided by mutation operators, which operate by random token and term exchanges. 
The number of terms to change has no strict limits. For tokens with parameters $(p_{k+1}, \; ... \; p_n) \in \mathbb{R}^{n-k}$, such as a parametric representation of an unknown external dependent variable, parameters are also optimized: the mutation is done with a random Gaussian increment. 



In order to combine structural elements of better equations, the cross-over operator is implemented. The interactions between parent equations are held on a term-level basis. 
The sets of terms pairs from the parent equation are divided into three groups: terms identical in both equations, terms that are present in both equations but have different parameters or only a few tokens inside of them are different, and the unique ones. The cross-over occurs for the two latter groups. For the second group it manifests as the parameter exchange between parents: the new parameters are selected from the interval between the parents' values. 

Cross-over between unique terms works as the complete exchange between them. The construction of exchange pairs between these tokens works entirely randomly. 


\subsection{Optimization of equation quality metric}
\label{subsec:single_obj}

The selection of the optimized functional distinguishes multiple approaches to the differential equation search. First of all, a more trivial optimization problem can be stated as in Eq.~\ref{eq:optimization_problem_operator}, where we assume the identity of the equation operator $F'(\overline{u}) = 0$ to zero as in Eq.~\ref{eq:data_driven_statement}. 

\begin{equation}
    Q_{op}(F'(u)) = || F'(\overline{u}) ||_n = || \sum_{i} \alpha_{i} \prod_{j} f_{ij} ||_n \longrightarrow \min_{\alpha_i \; t_{ij}}
\label{eq:optimization_problem_operator}
\end{equation}

An example of a more complex optimized functional is the norm of a discrepancy between the input values of the modelled variable and the solution proposed by the algorithm differential equation, estimated on the same grid. 
Classical solution techniques can not be applied here due to the inability of a user to introduce the partitioning of the processed domain, form finite-difference schema without a priori knowledge of an equation, proposed by evolutionary algorithm. An automatic solving method for candidate equation (viewed as in Eq.~\ref{eq:optimization_problem_solution_eq}) quality evaluation is introduced in \cite{maslyaev2022solver} to work around this issue. 

\begin{equation}
    Q_{sol}(F'(u)) = || u - \overline{u} ||_n \longrightarrow \min_{\alpha_i \; t_{ij}}
\label{eq:optimization_problem_solution}
\end{equation}

\begin{equation}
    F'(\overline{u}) = 0 : F'(\overline{u}) = \sum_{i} \alpha_{i} \prod_{j} f_{ij} = 0
\label{eq:optimization_problem_solution_eq}
\end{equation}

While both quality metrics Eq.~\ref{eq:optimization_problem_operator} and Eq.~\ref{eq:optimization_problem_solution} in ideal conditions provide decent convergence of the algorithm, in the case of the noisy data, the errors in derivative estimations can make differential operator discrepancy from the identity (as in problem in Eq.~\ref{eq:optimization_problem_operator}) an unreliable metric. Applying the automatic solving algorithm has high computational cost due to training a neural network to satisfy the discretized equation and boundary operators. 

As the single-objective optimization method for the study, we have employed a simple evolutionary algorithm with a strategy that minimizes one of the aforementioned quality objective functions. Due to the purposes of experiments on synthetic noiseless data, the discrepancy-based approach has been adopted. 

\subsection{Multi-objective optimization application}
\label{subsec:multi_obj}

As we stated earlier, in addition to process representation, the conciseness is also a valuable for regulating the interpretability of the model. Thus the metric of this property can be naturally introduced as Eq.~\ref{eq:norm_complexity}, with an adjustment of counting not the total number of active terms but the total number of tokens ($k_i$ for $i-th$ term).

\begin{equation}
     C(F'(u))= \#(F') = \sum_i k_i * \mathbf{1}_{\alpha_i \neq 0}
    \label{eq:norm_complexity}
\end{equation}

In addition to evaluating the quality of the proposed solution from the point of the equation simplicity, multi-objective enables the detection of systems of differential equations, optimizing qualities of modeling of each variable. 


While there are many evolutionary multi-objective optimization algorithms, MOEADD (Multi-objective evolutionary algorithm based on dominance and decomposition) \cite{moeadd} algorithm has proven to be an effective tool in applications of data-driven differential equations construction. We employ baseline version of the MOEADD from the aforementioned paper with the following parameters: PBI penalty factor $\theta = 1.0$, probability of parent selection inside the sector neighbourhood $\delta = 0.9$ ($4$ nearest sector are considered as ``neighbouring'') with $40\%$ of individuals selected as parents. Evolutionary operator parameters are: crossover rate (probability of affecting individual terms): $0.3$ and mutation rate of $0.6$.
The result of the algorithm is the set of equations, ranging from the most simplistic constructions (typically in forms of $\frac{\partial^{n} u}{\partial x_{k}^{n}} = 0$) to the highly complex equations, where extra terms probably represents the noise components of the dynamics. 




\section{Experimental study}
\label{sec:validation}

This section of the paper is dedicated to studying equation discovery framework properties. As the main object of interest, we designate the difference of derived equations between single- and multi-objective optimization launches. The validation was held on the synthetic datasets, where modelled dependent variable is obtained from solving an already known and studied equation. 

The tests were held on three cases: wave, Burgers and Korteweg-de Vries equations due to unique properties of each equation. The algorithms were tested in the following pattern: 64 evolutionary iterations for the single-objective optimization algorithm and 8 iterations of multi-objective optimization for the populations of 8 candidate equations, which resulted in roughly similar resource consumption.
10 independent runs are conducted with each setup. The main equation quality indicator in our study is the statistical analysis of the objective function mean ($\mu = \mu(Q(F'))$) and variance $\sigma^2 = (\sigma(Q(F')))^2$ among the different launches. 


The first equation was the wave equation as on Eq.~\ref{eq:burgers_equation} with the necessary boundary and initial conditions. The equation is solved with the Wolfram Mathematica software in the domain of $(x, t) \in [0, 1] \bigtimes [0, 1]$ on a grid of $101 \bigtimes 101$. Here, we have employed numerical differentiation procedures. 


\begin{equation}
\begin{aligned}
  \frac{\partial^2 u}{\partial t^2} &= 0.04 \frac{\partial^2 u}{\partial x^2} \\
\end{aligned}
\label{eq:burgers_equation}
\end{equation}

The algorithm's convergence due to the relatively simple structure was ensured in the case of both algorithms: the algorithm proposes the correct structure during the initialization or in the initial epochs of the optimization. However, such a trivial case can be a decent indicator of the ``ideal'' algorithm behaviour. The values of examined metrics for this experiment and for the next ones are presented on Tab.~\ref{tab:metrics}. 

\begin{table}[h!]
\caption{Results of the equation discovery}
\begin{tabular}{SSSSS} \toprule
    {metric}      & {method}           & {wave} & {Burgers} & {KdV} \\ \midrule
    {$\mu$}       & {single-objective} & {$5.72$} & {$2246.38$} & {$0.162$} \\
                  & {multi-objective}  & {$2.03$} & {$1.515$} & {$16.128$} \\
    {$\sigma^2$}  & {single-objective} & {$18.57$} & {$4.41*10^7$} & {$8.9 * 10^{-3}$} \\
                  & {multi-objective}  & {$0$}  & {$20.66$} & {$\approx 10^{-13}$}  \\ \bottomrule
\end{tabular}
\label{tab:metrics}
\end{table}

\begin{figure*}[!tbph]
    \subfigure[]{\includegraphics[width=0.27\textwidth]{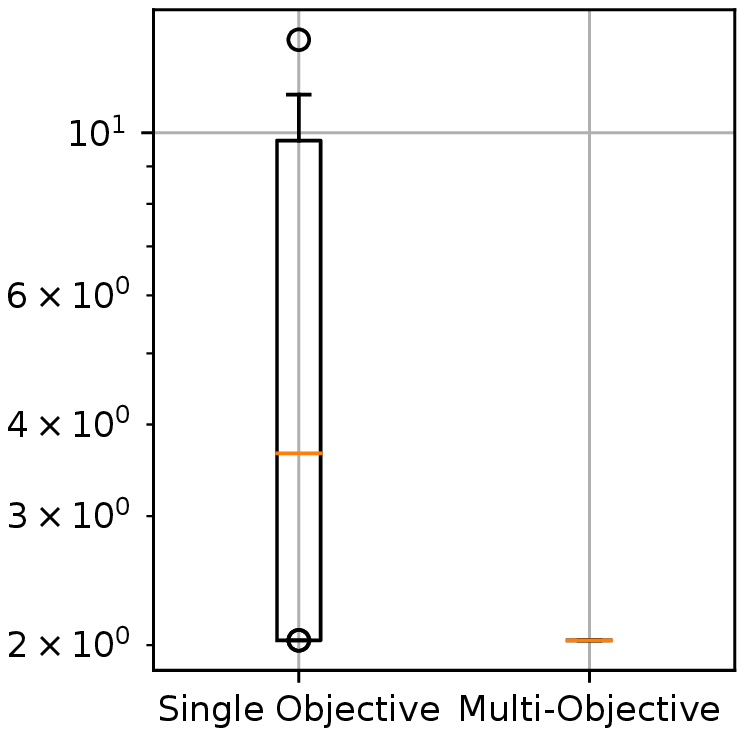}} 
    \subfigure[]{\includegraphics[width=0.27\textwidth]{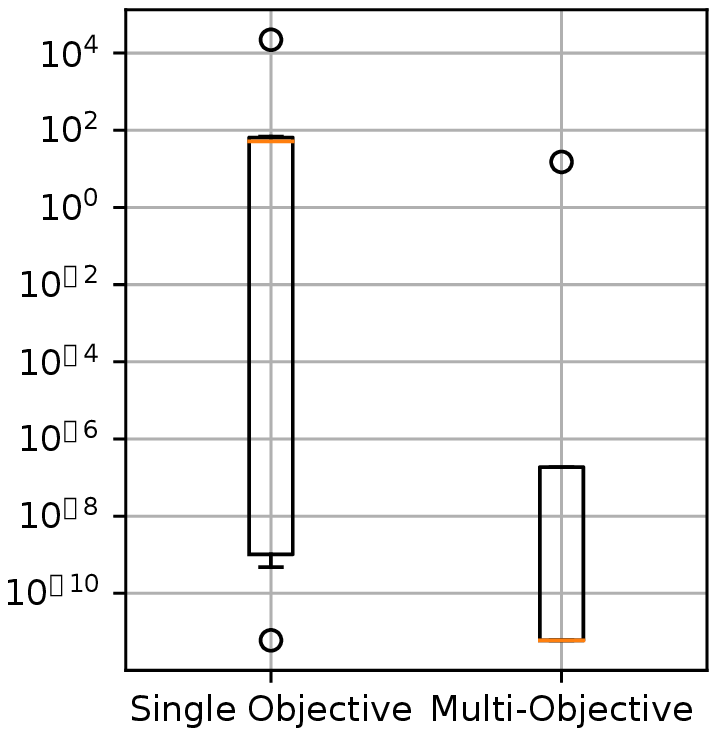}} 
    \subfigure[]{\includegraphics[width=0.27\textwidth]{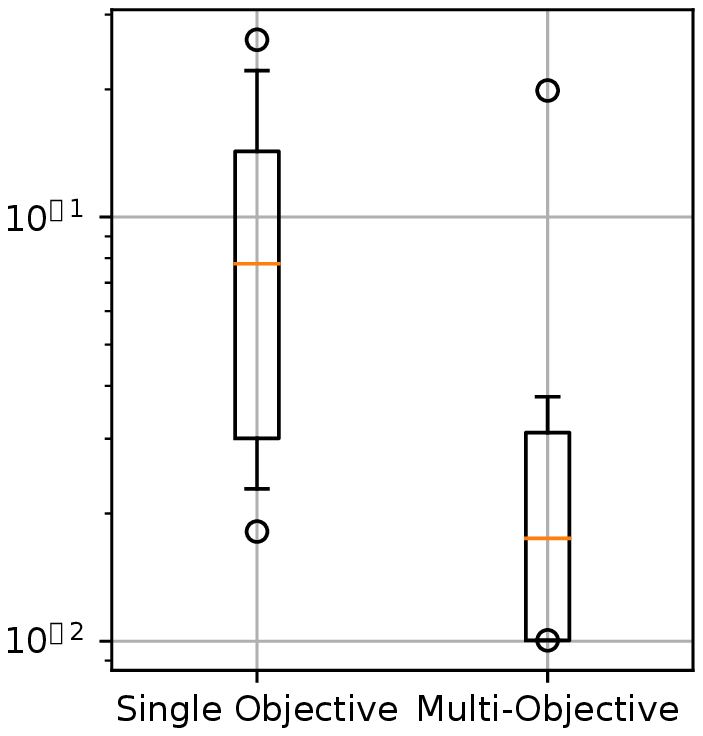}}
    \caption{Resulting quality objective function value, introduced as Eq.~\ref{eq:optimization_problem_solution_eq}, for single- and multi-objective approaches for (a) wave equation, (b) Burgers equation, and (c) Korteweg-de Vries equation}
    \label{fig:boxplots}
\end{figure*}


The statistical analysis of the algorithm performance on each equation is provided in Fig.~\ref{fig:boxplots}. 



Another examination was performed on the solution of Burgers' equation, which has a more complex, non-linear structure. The problem was set as in Eq.~\ref{eq:burgers}, for a case of a process without viscosity, thus omitting term $\nu \frac{\partial^2 u}{\partial t^2}$. As in the previous example, the equation was solved with the Wolfram Mathematica toolkit. 

\begin{equation}
\begin{aligned}
  \frac{\partial u}{\partial t} &+ u \frac{\partial u}{\partial x} = 0 \\
\end{aligned}
\label{eq:burgers}
\end{equation}

Derivatives used during the equation search were computed analytically due to the function not being constant only on small domain.


The presence of other structures that have relatively low optimized function values, such as $u'_{x} u'_{t} = u''_{tt}$, makes this case of data rather informative. Thus, the algorithm has a local optimum that is far from the correct structure from the point of error metric. 



The final set-up for an experiment was defined with a non-homogeneous Korteweg-de Vries equation, presented in Eq.~\ref{eq:KdV_equation}. The presence of external tokens in separate terms in the equation makes the search more difficult. 

\begin{equation}
\begin{aligned}
  \frac{\partial u}{\partial t} + 6 u \frac{\partial u}{\partial x} + \frac{\partial^3 u}{\partial x^3} = \cos{t} \sin{t}\\
\end{aligned}
\label{eq:KdV_equation}
\end{equation}

The experiment results indicate that the algorithm may detect the same equation in multiple forms. Each term of the equation may be chosen as the ``right-hand side'' one, and the numerical error with different coefficient sets can also vary. 


\section{Conclusion}
\label{sec:conclusion}

This paper examines the prospects of using multi-objective optimization for the data-driven discovery of partial differential equations. While initially introduced for handling problems of deriving systems of partial differential equations, the multi-objective view of the problem improves the overall quality of the algorithm. The improved convergence, provided by higher candidate individual diversity, makes the process more reliable in cases of equations with complex structures, as was shown in the examples of Burgers' and Korteweg-de Vries equations.

The previous studies have indicated the algorithm's reliability, converging to the correct equation, while this research has proposed a method of improving the rate at which the correct structures are identified. This property is valuable for real-world applications because incorporating large and complete datasets improves the noise resistance of the approach. 

The further development of the proposed method involves introducing techniques for incorporating expert knowledge into the search process. This concept can help generate preferable candidates or exclude infeasible ones even before costly coefficient calculation and fitness evaluation procedures.

\section{Code and Data availability}
\label{sec:repo}
The numerical solution data and the Python scripts, that reproduce the experiments, are available at the GitHub repository \footnote{https://github.com/ITMO-NSS-team/EPDE\_GECCO\_experiments}.

\section*{Acknowledgements}

This research is financially supported by the Ministry of Science and Higher Education, agreement FSER-2021-0012.

\bibliographystyle{ACM-Reference-Format}
\bibliography{main}

\appendix

\end{document}